# Traffic4cast – Large-scale Traffic Prediction using 3DResNet and Sparse-UNet

Bo Wang[1*], Reza Mohajerpoor[2], Chen Cai[2], Inhi Kim[3] and Hai L. Vu[1]

[1]*Institute of Transport Studies*
*Monash University, VIC 3800, Australia, Bo.Wang1@monash.edu*
[2]*CSIRO's Data61, NSW 2015, Australia*
[3]*Institute Civil and Environmental Engineering Department,*
*Kongju National University, South Korea*

## Abstract

The IARAI competition *Traffic4cast 2021*[1] aims to predict short-term city-wide high-resolution traffic states given the static and dynamic traffic information obtained previously. The aim is to build a machine learning model for predicting the normalized average traffic speed and flow of the subregions of multiple large-scale cities using historical data points. The model is supposed to be generic, in a way that it can be applied to new cities. By considering spatiotemporal feature learning and modeling efficiency, we explore 3DResNet and Sparse-UNet approaches for the tasks in this competition. The 3DResNet based models use 3D convolution to learn the spatiotemporal features and apply sequential convolutional layers to enhance the temporal relationship of the outputs. The Sparse-UNet model uses sparse convolutions as the backbone for spatiotemporal feature learning. Since the latter algorithm mainly focuses on non-zero data points of the inputs, it dramatically reduces the computation time, while maintaining a competitive accuracy. Our results show that both of the proposed models achieve much better performance than the baseline algorithms. The codes and pretrained models are available at https://github.com/resuly/Traffic4Cast-2021.

## 1 Introduction

Trac4cast competition at NeurIPS 2021, hosted by the Institute of Advanced Research in Artificial Intelligence (IARAI), is designed to solve the large-scale traffic prediction problem. Participants are required to predict the future traffic speed and volumes of grid cells in large-scale cities given historical information. This problem is crucial as several subsequent applications such as traffic navigation and traffic management require such information as the foundation for decision-making. However, city-scale traffic prediction is challenging because of complex and stochastic traffic movements over the spatiotemporal space.

---

1 See https://iarai.ac.at/traffic4cast/2021-competition/challenge/

In this competition, the dataset provided by HERE Technologies includes both static and dynamic traffic information of ten different cities worldwide. All given traffic data are aggregated every five minutes (one timestep), and the general forecasting problem can be formulated as follows:

With given input tensor $x = (T_h \times W \times H \times I)$ and static information $s = (W \times H)$, the model needs to predict target tensor $y = (T_f \times W \times H \times I)$, where $T_h = 12$ is the number of previous timesteps employed in the prediction model, $T_f = 6$ is the number of future timesteps to be predicted. $W = 495$ and $H = 436$ are the width and height of the city grid network, and $I = 8$ represents the traffic volume and speed values in four different directions. Moreover, $s$ describes the density of the road map in each cell of the city map.

Since the input and output data are very similar to image or video data format, several algorithms from computer vision are applicable to this problem. According to the previous competition reports (Kopp et al., 2021), the existing successful methods are mostly based on deep neural networks that have two main approaches for feature learning: (a) 2D convolution and (b) geometric convolution.

The first approach uses 2D convolution as the backbone for feature extraction. Many different structures such as U-Net (Choi, 2020; Martin, Hong, Bucher, Rupprecht, & Buffat, 2019) and HR-Net (Wu et al., 2020) have been proposed and achieved outstanding results. However, 2D convolution requires 3D data as inputs instead of the 4D data as in this problem. In practice, traffic features and temporal information are often combined to reduce dimensionality, but this operation may break the original spatiotemporal relationship from the input data space. Moreover, the temporal output is also a problem for the models mentioned above. The target requires 6 sequential timesteps which should be in a fixed temporal order, however it could not be strictly constrained by only one CNN output layer. Herruzo and Larriba-Pey (2020) have proposed ConvLSTM to enhance the temporal relationship, but this method is not computational efficient because all the convolutional operations are fed through the recurrent unit. To solve the above problems, we propose a novel 3DResNet with conditional CNN layers. To this end, the input data of the proposed scheme are treated as video and processed by 3D convolutional layers in the main feature learning process (see Figure 1). Moreover, we use additional 6 sequential CNN layers to generate the final output, where the previous layer's outputs are used as the next layer's inputs to restrain the temporal relationship. The results show that our proposed method can achieve the MSE of 50.22 on the core challenge test data (without averaging other models), against IARAI's baseline obtained from the best existing model.

On the other hand, geometric neural networks based methods treat all available road networks as a graph and perform graph convolutions, which can achieve better generalization for unseen traffic networks (Martin et al., 2020; Qi & Kwok, 2020). These methods learn the road network or traffic relationship more efficiently, as the refined data structure (graph) only contains the information of road connections. However, this method still maintains much redundant information because the traffic movement only happens on a partial graph in any given timestep. For example, 53.47% of cells of the Melbourne city map are ocean and thus never have any traffic. Roughly only 0.28% data points (on average) are non-zero with given $x$ and $y$ pairs in one day. The highly sparse distribution of the traffic data results in overwhelmed null operations on zero regions in both graph and traditional convolutions. We have introduced a new variant of U-Net using sparse convolution (Graham et al., 2018) as the backbone of feature extraction to solve this shortcoming. This method dynamically converts the non-zero data points into a list and only performs high-dimensional convolution on non-zero state regions. Our results pinpoint that sparse convolution can improve the inference speed more than six times with the same model structure and device. In addition, it achieved competitive results on the extended challenge.

## 2   Data Preparation

There are data of ten different cities in the entire dataset. The static data are offered for all cities, and the availability of dynamic traffic data is different due to the challenge settings. Four cities (Moscow,

Bangkok, Antwerp, and Barcelona) have complete training data in both 2019 and 2020. Another four cities (Melbourne, Chicago, Istanbul, and Berlin) for the core challenge only have data in 2019 and They need to predict the test set in 2020. Moreover, the extended challenge comprises two cities (New York and Vienna), which pertain to no training data and we need to predict the test dataset for both 2019 and 2020 years.

The competition in 2021 aims to predict the traffic for unseen times (pandemic periods in 2020) or locations (New York and Vienna that have no training data). Since the complete testing samples are not accessible for all these unseen scenarios, there is no chance to fine-tune the trained model as few-shot learning or transfer learning. J. Wang et al. (2021) summarize this type of task as *domain generalization,* which aims to train a model that has enough generalization ability for a new domain. Many different strategies have been proposed in this area, and this paper mainly explored the approach of data generation that provides diverse samples to help model generalization.

We have designed a custom data loading scheme to enhance data diversity and loading speed in each training epoch. The original plan provided by the host randomly shuffles the training data, that can help the training batches to represent the overall data distribution and lead to better performance. However, since the fact that the data are stored day by day (in h5py format), the process of file loading becomes time-consuming as each batch contains indices from multiple days.

Instead of shuffling all indices across all files, our scheme prepares the training samples in two stages. First, we sample the files without a replacement that contains all different cities, years, and days of the week as a training epoch to remain a consistent sample diversity. Second, the 240 indices inside each randomly ordered file will be locally shuffled. With the batch size of 2 and 2 CPU workers, the pure data loading speed increased from 1.89 batches/s up to 60 batches/s. Accordingly, the waiting time for GPUs has been dramatically reduced by this process.

## 3 Proposed Models

### 3.1 3DResNet

To perform the convolutional operation through the entire spatial and temporal space across the input data, we adopt 3D convolution (Tran, Bourdev, Fergus, Torresani, & Paluri, 2015; Wang, Vu, Kim, & Cai, 2021) for the input tensor $x = (T_h \times W \times H \times I)$. Note that this paper only uses the historical traffic data as the model input and no particular feature engineering method is applied.

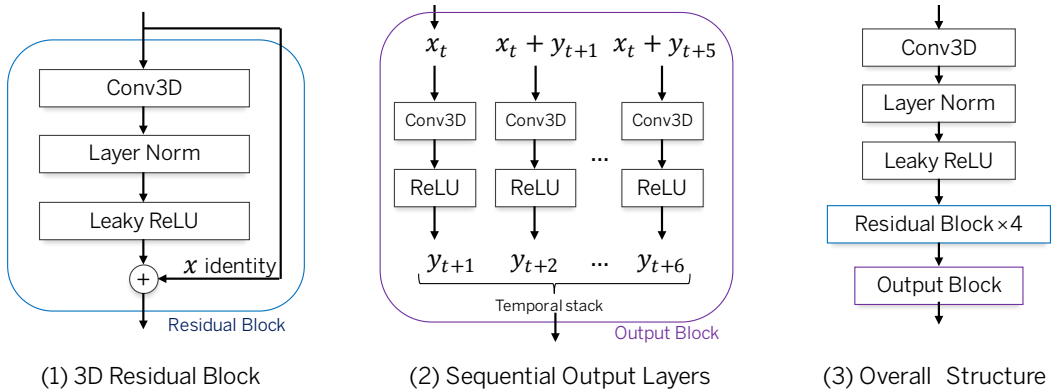

(1) 3D Residual Block  (2) Sequential Output Layers  (3) Overall Structure

Figure 1: Model structure of 3DResNet. All convolutional layers use kernel size of 3, padding 1, and stride 1 to keep the size of the study area.

Figure 1 shows the overall structure of the proposed 3DResNet. We use the first Conv3D layer in subplot (3) to increase the hidden size from 12 to 16 for the following residual layers. Then the residual blocks are used for increasing the model depth. Finally, since the results of $y_t$ to $y_{t+1}$ are dependent on previous output, the entire output block produces the prediction sequentially for each future timestep.

In practice, we first replace the output block in Figure 1 with a regular Conv3D layer for a warm-up training, then set up the complete structure for further optimization. We found this approach can make the model easier to converge.

3.2 Sparse-UNet

As introduced in Section 1, GCN (geometric neural networks) or CNN (convolutional neural networks) needs many extra computations when dealing with sparse inputs. To increase the efficiency of data processing and modeling, the dense tensors can be converted into sparse tensors (Graham, 2014). Coordinate list (COO) is the most common format that uses a list of coordinates and features to represent the sparse tensor. In this paper, we proposed a simple UNet structure using sparse convolution based on COO (Choy, Gwak, & Savarese, 2019; Tang et al., 2020), see Figure 2.

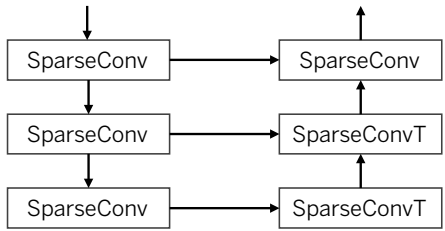

Figure 2: Model structure of Sparse-UNet. All convolutional layers use kernel size of 3, padding 1, and stride 1 to keep the target size. ConvT means the transposed sparse convolution.

## 4 Results

Models proposed in this paper were trained with two P100 GPUs. Through our experiments and the submission results, we found a strong positive relationship between training MSE and testing MSE. Therefore, we use all data for training and select the lowest training MSE as the best model.

As a result, the 3DResNet achieved 50.23 MSE for the core challenge (UNet baseline provided by IARAI is 51.28), and Sparse-UNet achieved 61.59 MSE for the extended challenge (Naive average baseline provided by IARAI is 63.14). The following section further discusses our findings through the training experiments.

## 5 Discussion

### 5.1 3DResNet

To understand the impact of different components proposed in section 3.1, we further implement two variants of 3DResNet for comparison. There are three different model types explained as follows, and their training performance is displayed in Figure 3.

- *Conv3D+SequentialOutput* represents the full version of the proposed model.
- *Conv3D+ConvOutput* uses one layer of Conv3D to replace the *SequentialOutput* block.
- *Conv2D+ConvOutput* reduces the dimension by stacking the timesteps and traffic features into the same axis and applies Conv2D for all convolutional layers.

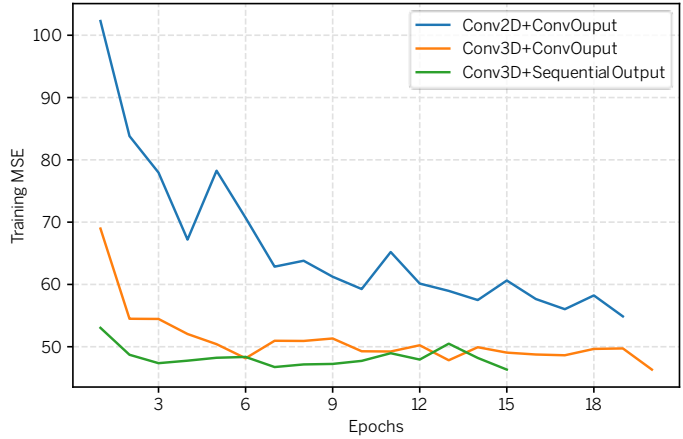

Figure 3: Training performance of different ResNet structures. MSE represents the average loss of all different cities described in Section 2.

The training trends in Figure 3 show that: (1) with the same settings of model structure and hidden size, the 2D CNN converges much slower compared to 3D CNN models, and (2) by applying the *SequentialOutput* block, the training MSE is more stable and lower.

## 5.2 Sparse Convolution

This section further discusses the impact of using sparse convolution. On top of the existing Sparse-UNet, we have also implemented a regular Conv3D-UNet for comparison. The latter has the identical parameter setting but uses Conv3D for convolutional layers.

Table 1. Different training times using Conv3D and sparse convolution (batch size = 2 with one P100 GPU). City names from left to right: Antwerp, Bangkok, Barcelona, Berlin, Chicago, Istanbul, Melbourne, Moscow.

| City | ANT | BAN | BAR | BER | CHI | IST | MEL | MOS |
| --- | --- | --- | --- | --- | --- | --- | --- | --- |
| Non-zero Rate (per batch) | 0.0079 | 0.0072 | 0.0023 | 0.0303 | 0.0085 | 0.0481 | 0.0039 | 0.0758 |
| Conv3D-UNet (batches/s) | 0.98 | 0.98 | 0.98 | 0.98 | 0.98 | 0.98 | 0.98 | 0.98 |
| Sparse-UNet (batches/s) | 7.76 | 6.82 | 8.30 | 5.56 | 7.87 | 3.53 | 8.45 | 2.18 |

The training speed improvement using sparse convolution on different samples can be found in Table 1. Since the Conv3D-UNet performs convolution operation on all input tensors, it remains the same processing speed for all different samples. In contrast, Sparse-UNet increases the processing speed on average by 6.2 times. Particularly, the speed varies with varying proportions of non-zero values in different cities.

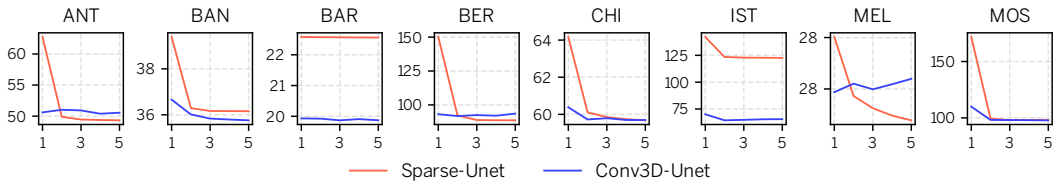

Figure 4: The different training performances of Sparse-UNet and Conv3D-UNet.

Although Sparse-UNet has better training speed, its representation learning ability changes with different samples. We have trained a small 3-days sample from each city for five epochs from starch using Conv3D-UNet and Sparse-UNet, respectively. Their changes of training MSE are displayed in Figure 4, which reflects the following findings:

- The convergence speeds are different for each city. There is no dominant method that can handle all different data distributions.

- The ranges of MSE are very different. Mostly, the city that has lower sparsity would lead to a lower MSE as well.

From our submission experience, the traditional Conv-based methods perform well on the core challenge, but they are worse than sparse-based models on the extended challenge. Conversely, the sparse-based method can be generalized well on the extended challenge but performs worse on the core challenge. This phenomenon may be caused by the specific model structure design or the setting of hyper-parameters. Nevertheless, sparse convolution's full performance and features still have not been fully explored with the regression problem. More complex model design or tuning methods can be reached by neural architecture search (NAS) and parameter optimization methods in the future.

### 5.3 Performance Bottleneck

This section analyzes the bottleneck of current forecasting models through forecasting visualizations. The forecasting problem discussed below has been found in both Conv-based and Sparse-based models in different cities. We use a forecasting snippet in Figure 5 to demonstrate the common bottleneck of this task.

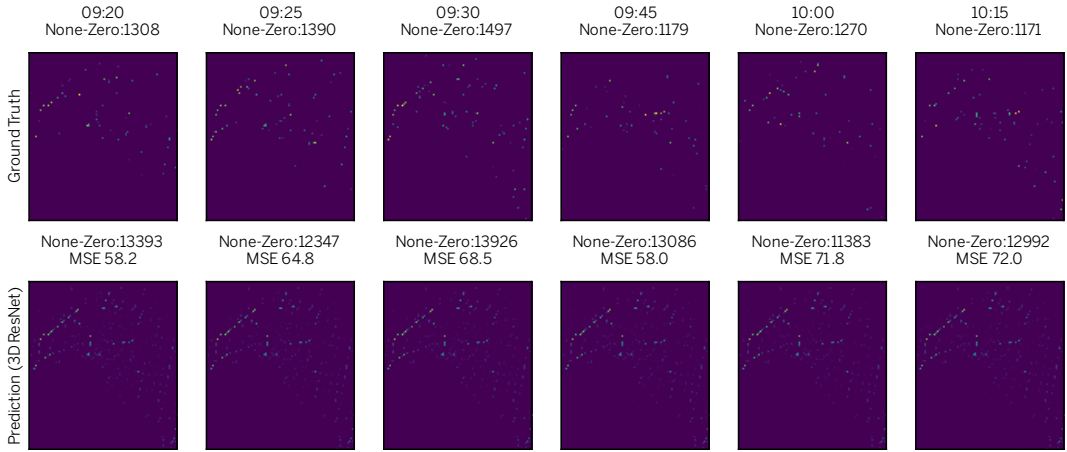

Figure 5: The northeast traffic speed prediction in six sequential timesteps. The model used here is our best-trained 3DResNet, and the target data is on 2019-01-08, Melbourne city. None-Zero: the number of non-zero data points in each counterpart 495×436 grid.

In general, the forecasting targets are extremely sparse, see Figure 5. In each timestep, the forecasting model needs to decide about 1,500 non-zero positions through a map with 215,820 spaces. This task is challenging because the previous traffic flow could be spatially decreased or increased in many ways. The current model still has room to be further improved in spatial granularity.

In terms of forecasting results, the model produced a similar traffic distribution for all timesteps, which indicates that the model is looking for the average traffic of the next six timesteps. Although our designed *SequentialOutput* block enhanced the temporal relationship by changing the input sequence, the prediction did not clearly change temporally. This phenomenon was caused by the data sparsity and the loss function. According to our observations, two specific traffic distributions in five minutes from different days can be very different even they are the same city and at the same time. This makes spatial prediction especially difficult for the existing models. Therefore, our trained model tends to produce many small values across the road network to reduce the average MSE. However, once this strategy or optimization direction has been formed, then the model would be stuck in a local minimum only to produce an average distribution.

Lastly, it is hard to show the specific or significant traffic changes in each feature timestep from the forecasting results. The forecasting models still need further improvement to produce meaningful implications for local traffic management.

## 6 Conclusion

This paper presented the 3DResNet and Sparse-UNet approaches for large-scale traffic prediction tasks in *traffic4cast* competition. These two methods have achieved competitive results in the core and extended challenges, respectively. Furthermore, our results showed that (1) the 3D convolution can achieve better performance than 2D convolution with the same model structure, (2) our proposed sequential output block can stabilize the model training with a lower MSE, and (3) the sparse convolution can improve the training speed significantly (6x than Conv3D) while achieving a competitive generalization result.

A few interesting directions can be further investigated in future works. Firstly, the study of sparse convolution for the regression problem is still in an early stage. The inner mechanism, structure design, and model generalization of this method need further investigation. Secondly, the different approaches of convolution could perform differently with various samples. Future research could consider the strength of the different types of convolutions and make appropriate combinations. Thirdly, there are some early studies in using AI for discovering laws of physics by approximating differential equations. This direction could also be applied to the kinetic traffic flow model and further improve the transfer learning from one transport network to another. Finally, the current results are still limited due to the temporal repeating problem. More metrics and loss functions should be proposed to improve the interpretability of city-scale traffic forecasting.